\documentclass[11pt,a4paper]{article}
\usepackage[hyperref]{acl2020}

\usepackage{textcomp} %N % for apostrophe
\usepackage{times}
\usepackage{latexsym}
\usepackage{multirow}
\usepackage{amsmath}
\usepackage{graphicx} % for inserting the image

\usepackage{booktabs}

\usepackage{algorithm} %N
\usepackage[noend]{algpseudocode} %N

\makeatletter
\def\algbackskip{\hskip-\ALG@thistlm}
\makeatother
\usepackage{xcolor}
\usepackage{soul}

\newcommand{\hlc}[2][yellow]{{%
    \colorlet{foo}{#1}%
    \sethlcolor{foo}\hl{#2}}%
}
\usepackage{array}

\newcolumntype{M}[1]{>{\centering\let\newline\\\arraybackslash\hspace{0pt}}m{#1}}
\newcolumntype{L}[1]{>{\centering\let\newline\\\arraybackslash\hspace{0pt}}m{#1}}
\newcolumntype{C}[1]{>{\centering\let\newline\\\arraybackslash\hspace{0pt}}m{#1}}
\newcolumntype{R}[1]{>{\centering\let\newline\\\arraybackslash\hspace{0pt}}m{#1}}
\newcolumntype{P}[1]{>{\centering\let\newline\\\arraybackslash\hspace{0pt}}m{#1}}

\usepackage{microtype}

\aclfinalcopy % Uncomment this line for the final submission
\def\aclpaperid{1549} %  Enter the acl Paper ID here

\title{Syn-QG: Syntactic and Shallow Semantic Rules for Question Generation}

\author{Kaustubh D. Dhole \\
  Amelia Science \\
  R\&D, IPsoft \\
  New York, NY 10004 \\
  \texttt{kdhole@ipsoft.com} \\\And
  Christopher D. Manning \\
  Department of Computer Science \\
  Stanford University  \\
  Stanford, CA 94305 \\
  \texttt{manning@stanford.edu} \\}

\date{}

\begin{document}
\maketitle
\begin{abstract}
 Question Generation (QG) is fundamentally a simple syntactic transformation; however, many aspects of semantics influence what questions are good to form. We implement this observation by developing Syn-QG, a set of transparent syntactic rules leveraging universal dependencies, shallow semantic parsing, lexical resources, and custom rules which transform declarative sentences into question-answer pairs. We utilize PropBank argument descriptions and VerbNet state predicates to incorporate shallow semantic content, which helps generate questions of a descriptive nature and produce inferential and semantically richer questions than existing systems. In order to improve syntactic fluency and eliminate grammatically incorrect questions, we employ back-translation over the output of these syntactic rules. A set of crowd-sourced evaluations shows that our system can generate a larger number of highly grammatical and relevant questions than previous QG systems and that back-translation drastically improves grammaticality at a slight cost of generating irrelevant questions.
\end{abstract}

\section{Introduction}

Automatic Question Generation (QG) is the task of generating question-answer pairs from a declarative sentence. It has direct use in education and generating engagement, where a system automatically generates questions about passages that someone has read. A more recent secondary use is for automatic generation of questions as a data augmentation approach for training Question Answering (QA) systems. QG was initially approached by syntactic rules for question-generation, followed by some form of statistical ranking of goodness, e.g., ~\cite{heilmanquestion, heilman2010good}. In recent years, as in most areas of NLP, the dominant approach has been neural network generation~\cite{du2017learning}, in particular using a sequence-to-sequence architecture, which exploits the data in the rapidly growing number of large QA data sets.
\begin{figure}
\begin{center}
  \includegraphics[width=\linewidth]{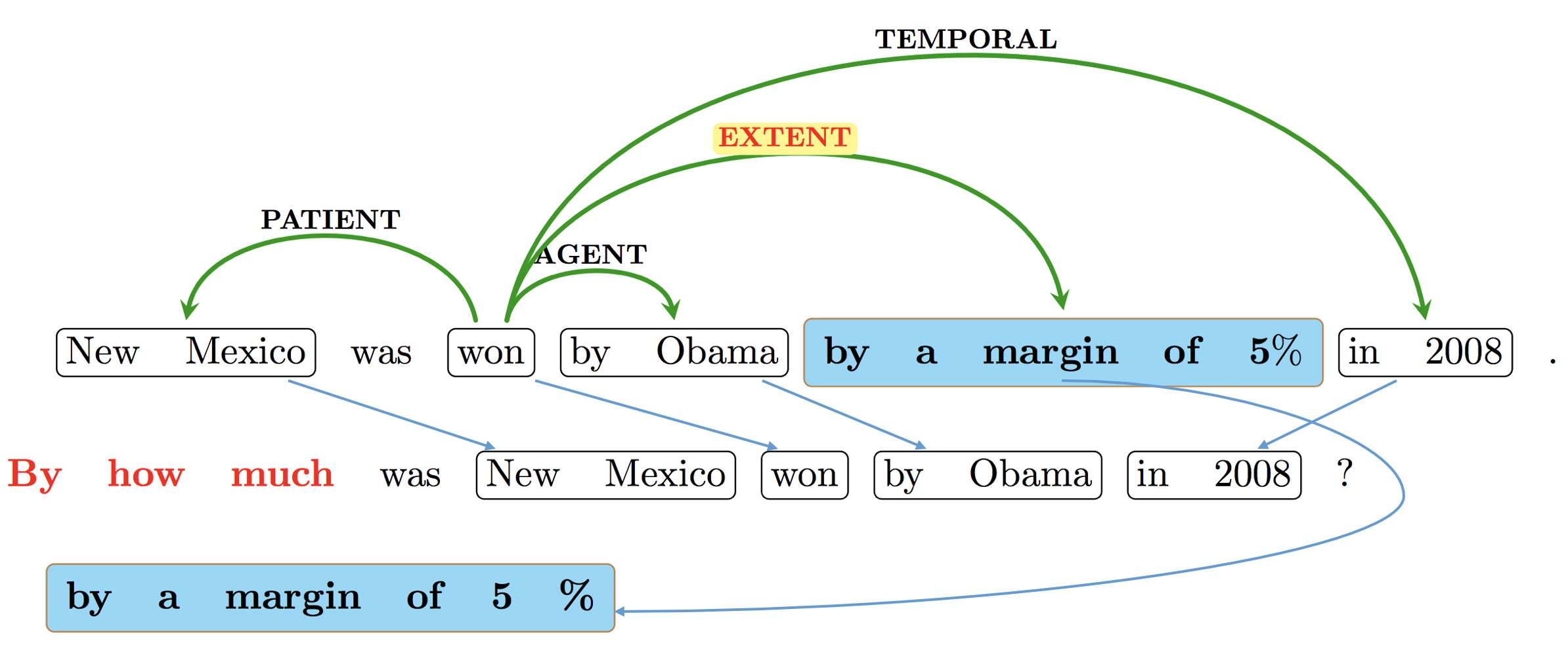}
  \end{center}
  \caption{ The SRL structure is leveraged to invoke a template, and a simple rearrangement of the modifying arguments is performed.}
  \label{fig:srl_qgen}
\end{figure}

Previous rule-based approaches suffer from a significant lack of variety in the questions they generate, sticking to a few simple and reliable syntactic transformation patterns. Neural architectures provide a pathway to solving this limitation since they can exploit QA datasets to learn the broad array of human question types, providing the usual neural network advantages of a data-exploiting, end-to-end trainable architecture. Nevertheless, we observe that the quality of current neural QG systems is still lacking: The generated questions lack syntactic fluency, and the models lack transparency and an easy way to improve them.

\begin{figure}[!ht]
\begin{center}
  \includegraphics[scale= 0.15]{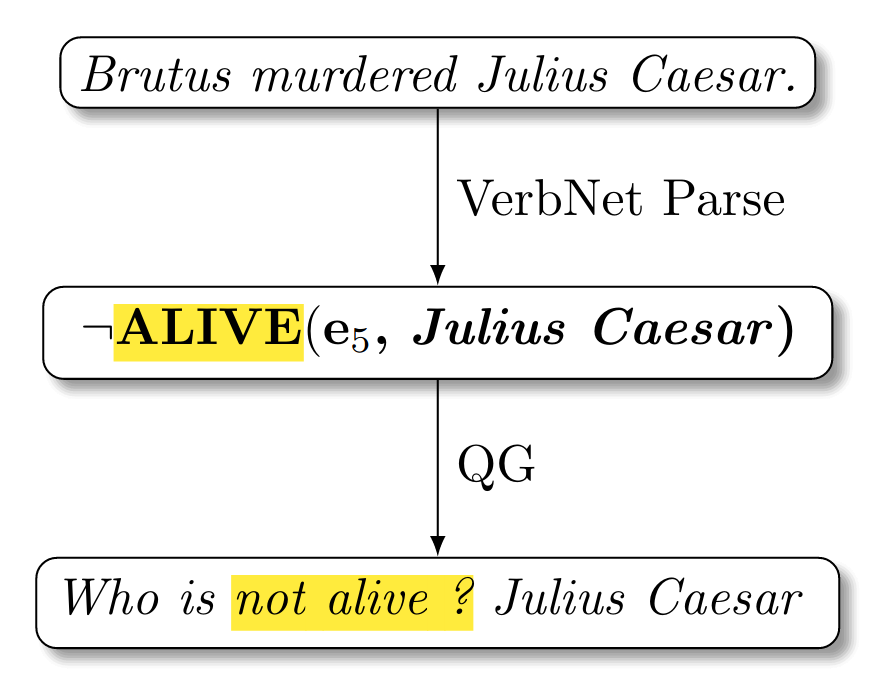}
  \end{center}
  \caption{VerbNet Predicate Question Generation. Detailed intermediate steps are described in Figure~\ref{fig:vn1}.}
  \label{fig:vn4}
\end{figure}

We argue that in essence QG can be governed by simple syntactic ``question transformations'' -- while the implementation details vary, this is in accord with all major linguistic viewpoints, such as Construction Grammar and Chomskyan Generative Grammar, which emphasize grammatical rules and the existence of finite ways to create novel utterances. However, successful, fluent question generation requires more than just understanding syntactic question transformations, since felicitous questions must also observe various semantic and pragmatic constraints. We approach these by making use of semantic role labelers (SRL), previously unexploited linguistic semantic resources like VerbNet's predicates (Figure~\ref{fig:vn4}) and PropBank's rolesets and custom rules like implications, allowing us to generate a broader range of questions of a descriptive and inferential nature. A simple transformation commonly used in rule-based QG is also displayed in Figure~\ref{fig:srl_qgen}.

We evaluate our QG framework, Syn-QG against three QG systems on a mixture of Wikipedia and commercial text sentences outperforming existing approaches in grammaticality and relevance in a crowd-sourced human evaluation while simultaneously generating more types of questions. We also notice that back-translated questions are grammatically superior but are sometimes slightly irrelevant as compared to their original counterparts. The Java code is publicly available at ~\href{https://bitbucket.org/kaustubhdhole/syn-qg/}{https://bitbucket.org/kaustubhdhole/syn-qg/}.

\section{Related Work}
With the advent of large-scale QA datasets~\cite{rajpurkar2016squad,nguyen2016ms}, recent work in QG~\cite{du2017learning,zhou2017neural} has primarily focused on training sequence-to-sequence and attention-based architectures.~\citet{dong2019unified} fine-tuned the question generation task by taking advantage of a large pre-trained language model. Success in reinforcement learning has inspired teacher-student frameworks~\cite{wang2017joint,tang2017question} treating QA and QG as complementary tasks and performing joint training by using results from QA as rewards for the QG task.~\citet{yuan-etal-2017-machine, Hosking_2019, zhang2019addressing} used evaluation metrics like BLEU, sentence perplexity, and QA probability as rewards for dealing with exposure bias.

~\citet{chen2019natural} trained a reinforcement learning based graph-to-sequence architecture by embedding the passage via a novel gated bi-directional graph neural network and generating the question via a recurrent neural network. To estimate the positions of copied words,~\citet{liu2019learning} used a graph convolution network and convolved over the nodes of the dependency parse of the passage.~\citet{li2019improving} jointly modeled OpenIE relations along with the passage using a gated-attention mechanism and a dual copy mechanism.
 
Traditionally, question generation has been tackled by numerous rule-based approaches~\cite{heilmanquestion,mostow2009generating,yao2010question,lindberg2013generating,labutov2015deep}.~\citet{heilmanquestion, heilman2010good} introduced an overgenerate-and-rank approach that generated multiple questions via rule-based tree transformations of the constituency parse of a declarative sentence and then ranked them using a logistic-regression ranker with manually designed features.~\citet{yao2010question} described transformations of Minimal Recursion Semantics representations guaranteeing grammaticality. Other transformations have been in the past defined in terms of templates~\cite{mazidi2014linguistic, mazidi2015leveraging, mazidi2016infusing, flor2018semantic}, or explicitly performed~\cite{heilmanquestion} by searching tree patterns via Tregex, followed by their manipulation using Tsurgeon~\cite{levy2006tregex}.~\citet{kurdi2020systematic} provide a comprehensive summary of QG, analysing and comparing approaches before and after 2014.

Vis-\`{a}-vis current neural question generators, rule-based architectures are highly transparent, easily extensible, and generate well-formed questions since they perform clearly defined syntactic transformations like subject-auxiliary inversion and \textit{WH-movement} over parse structures whilst leveraging fundamental NLP annotations like named entities, co-reference, temporal entities, etc. 

However, most of the existing rule-based systems have lacked diversity, being mostly focused on generating \textit{What}-type and boolean questions and have mainly exploited parse structures which are not semantically informed.~\citet{mazidi2016infusing,flor2018semantic} use Dependency, SRL, and NER templates but do not handle modalities and negation in a robust manner. Moreover, there is plenty of availability of core linguistic resources like VerbNet and PropBank, which provide further unique ways to look at sentences and ask questions differently besides the generally well-established dependency and SRL parses.

\section{Syn-QG}
Syn-QG is a rule-based framework which generates questions by identifying potential short answers in 1) the nodes of crucial dependency relations 2) the modifying arguments of each predicate in the form of semantic roles 3) named entities and other generic entities 4) the states of VerbNet's thematic roles in the form of semantic predicates and 5) PropBank roleset specific natural language descriptions. Each of the five heuristics works independently, generating a combined set of question-answer pairs, which are eventually back-translated. We describe each of these five sources.

\subsection{Dependency Heuristics}

Dependency trees are syntactic tree structures, wherein syntactic units in the form of words are connected via directed links. The finite verb is considered as the structural root of the tree, and all other syntactic units are either directly (\textit{nsubj, dobj,}, etc.) or indirectly (\textit{xcomp, iobj,} etc.) dependent on this finite verb.

We present rules over such dependency trees annotated according to the Universal Dependencies (UD) format~\cite{de-marneffe-etal-2014-universal}. To extract dependency structures, we use the parser of~\citet{gardner2018allennlp}.

We make use of PropBank\textquotesingle s predicate-argument structure (SRL) for clausal extraction of the verb headed by a select few dependency nodes which can serve as answers. These rules treat the clause as a combination of a subject, an object, the head verb and other non-core arguments. The clause is further refined with modals, auxiliaries and negations if found around the verb. Finally, we make use of a set of predefined handwritten templates, a few of which are described in Table~\ref{dep-table}. 
\begin{table*}[!hbpt]
\centering
\small
\begin{tabular}{| M{2cm} | L{3.1cm} | C{5cm} | R{4.2cm} |}
\hline
\textbf{Potential Short Answer} \newline \textbf{(Dependencies)} &
\textbf{Question Template} &
\textbf{Sample Fact} &
\textbf{Generated Question} \\
\hline
& & & \\
subject (\hlc[cyan!40]{nsubj}) &
\textbf{Wh} mainAux otherAux \hlc[pink]{verb} \hlc[yellow!40]{obj} \hlc[green!40]{modifiers}? &
\hlc[cyan!40]{Ricky Ponting} accepted captaincy during Australia's golden era.  &
\textbf{Who} \hlc[pink]{accepted} \hlc[yellow!40]{captaincy} \hlc[green!40]{during Australia's golden era}? \\
 & & & 
\\
direct object(\hlc[cyan!40]{dobj}) & \textbf{Wh} \hlc[cyan!20]{mainAux} \hlc[yellow!40]{nsubj} otherAux \hlc[pink]{verb} \hlc[green!40]{modifiers}? &
 In monsoon, India receives \hlc[cyan!40]{large amounts of rain that can cause flooding}. & \textbf{What} \hlc[cyan!20]{does} \hlc[yellow!40]{India} \hlc[pink]{receive} \hlc[green!40]{in monsoon}?     \\
& & & 
\\
open clausal complement (\hlc[cyan!40]{xcomp})    &
\textbf{Wh} \hlc[cyan!20]{mainAux} \hlc[yellow!40]{nsubj} \hlc[pink]{verb} \hlc[green!40]{modifiers}?    &    
The Sheriff did not try \hlc[cyan!40]{to eat the apples} while the outlaws were fasting. &
\textbf{What} \hlc[cyan!20]{did} \hlc[yellow!40]{the Sheriff} \hlc[pink]{not try} \hlc[green!40]{while the outlaws were fasting}? \\
& & & 
\\
copula (\hlc[cyan!40]{cop}) &
\textbf{How would you describe} \hlc[yellow!40]{nsubj}? &
Comets are \hlc[cyan!40]{leftovers from the creation of our solar system about 4.5 billion years ago}. &
\textbf{How would you describe} \hlc[yellow!40]{comets} ? \\

\hline
\end{tabular}
\caption{A few templates to describe the construction of questions. Different word units are shown in unique colors to describe the filling of the template. All the short answers are highlighted in \hlc[cyan!40]{blue}.}
\label{dep-table}
\end{table*}

In each of the templates, we convert \emph{What} to \emph{Who/Whom}, \emph{When} or \emph{Where} depending on the named entity of the potential answer and \emph{do} to \emph{does} or \emph{did} according to the tense and number of the subject to ensure subject-verb agreement.
%\\\textbf{do} = \{\emph{do,does,did}\} \\
%\textbf{What} = \{\emph{What,Who, Whom}\}\\
The pseudo code is described in Algorithm~\ref{alg:the_alg} of the Appendix.

\subsection{SRL Heuristics}

While dependency representations are perhaps the most popular syntactic method for automatically extracting relationships between words, they lack sufficient semantic detail. Being able to answer \emph{``Who} did \emph{what} to \emph{whom} and \emph{how, why, when} and \emph{where''} has been a central focus in understanding language. In recent decades, shallow semantic parsing has been a prominent choice in understanding these relationships and has been extensively used in question generation~\cite{mazidi2016infusing, flor2018semantic}. 

PropBank-style frames provide semantically motivated roles that arguments around a verb play. Moreover, highly accurate semantic role labeling models are being developed owing to corpora like PropBank and FrameNet. We take advantage of the SRL model of~\citet{gardner2018allennlp} for extracting the roles of each verb in the sentence. %Specifically, viewing a sentence in terms of its predicate-argument structures eliminates the need for a separate sentence simplification model.

\begin{algorithm}[!hbpt]
\caption{SRL Heuristics}\label{srl-algo}
\begin{algorithmic}
\State $\{SRL_1 \ldots SRL_s \} \gets SRL(w_0 \ldots w_n)$
\BState \emph{loop j = 0, until j = s}:
\If {$SRL_j \; contains \:A_0 \text{ or } A_1\: and\: at\: least\: 1\: A_m$}
\State $\{A_0 \ldots A_{CAU}, A_{TMP}\} \gets SRL_j$
\BState \emph{$loop \text{ } A_x \in SRL_j \; if \; A_x = modifier$}:
\State $subj \gets A_0$
\State $A_{x}^{-} \gets \sum (A_3,A_4,... A_{TMP} - A_x)$
\State $verb \gets \{A_v, modals, negation\}$
\State $template \gets modifier_{type} \gets A_x$
\State $QA \gets template(subj, A_x, verb, A_{x}^{-}) $
\State \textbf{close};
\EndIf
\end{algorithmic}
\end{algorithm}

We succinctly describe the steps taken in Algorithm~\ref{srl-algo}. We first filter out all the predicates which have an \emph{Agent} or a \emph{Patient} and at least one other modifier like \emph{Extent, Manner, Direction,} etc. These modifiers would serve as our short answers. We make use of a set of predefined handwritten templates described in Table~\ref{srl-table}, which rearrange the arguments within the fact to convert it into an interrogative statement depending on the modifier.  

In Figure~\ref{fig:srl_qgen}, the predicate ``won'' is modified by a \emph{Patient} ``New Mexico'', an \emph{Agent} ``Obama'', an \emph{Extent} modifier ``by a margin of 5\%'' and a \emph{Temporal} modifier ``in 2008''. For \emph{Extent} as a short answer, we fill a pre-defined template ``By how much mainAux nsubj otherAux verb obj modifiers ?'' to get the above question-answer pair. We keep the order of arguments as they appear in the original sentence. The templates are described in Table~\ref{srl-table}.

\subsection{Named Entities, Custom Entities, and Hypernyms}
We create separate templates when any numbered SRL argument contains common named entities like \textit{Person}, \textit{Location}, \textit{Organization} etc. Like~\citet{flor2018semantic}, we add specific rules in the form of regexes to address special cases to differentiate between phrases like \emph{\textbf{For how long}} and \emph{\textbf{Till when}} instead of a generic \emph{\textbf{When}} question type. Some of the templates are described in Table~\ref{ner-table} in the Appendix. The approach is described in Algorithm~\ref{ner-algo} in the Appendix.

\begin{table*}[ht!]
\small
\centering
\begin{tabular}{| M{2.3cm} | L{3cm} | C{5cm} | R{4cm} |}
\hline
\textbf{Potential Short Answer} \newline \textbf{(Verb Arguments)} &
\textbf{Question Template} &
\textbf{Sample Fact} &
\textbf{Generated Question} \\
\hline
& & & \\
Locative (\hlc[cyan!40]{LOC}) &
\textbf{Where} \hlc[cyan!20]{mainAux} \hlc[yellow!40]{nsubj} otherAux \hlc[pink]{verb} \hlc[yellow!40]{obj} \hlc[green!40]{modifiers} ? &
Americans eat about 100 acres of pizza each day, with about 3 billion pizzas sold annually \hlc[cyan!40]{in the USA.} & 
\textbf{Where} \hlc[cyan!20]{do} \hlc[yellow!40]{about 3 billion pizzas} \hlc[pink]{sell} \hlc[green!40]{annually} ?
\\
& & & \\
Manner  (\hlc[cyan!40]{MNR}) &
\textbf{How} \hlc[cyan!20]{mainAux} \hlc[yellow!40]{nsubj} otherAux \hlc[pink]{verb} \hlc[yellow!40]{obj} \hlc[green!40]{modifiers} ? &
Young Sheldon was caught unaware as the liquid was oozing out of the chamber \hlc[cyan!40]{in a zig-zag fashion.} &
\textbf{How} \hlc[cyan!20]{was} \hlc[yellow!40]{the liquid} \hlc[pink]{oozing}  \hlc[green!40]{out of the chamber}?  \\
Purpose (\hlc[cyan!40]{PNC} and \hlc[cyan!40]{PRP}) &
\textbf{For what purpose} \hlc[cyan!20]{mainAux} \hlc[yellow!40]{nsubj} otherAux \hlc[pink]{verb} \hlc[yellow!40]{obj} \hlc[green!40]{modifiers} ? &
Collectively, South African women and children walk a daily distance equivalent to 16 trips to the moon and back \hlc[cyan!40]{to fetch water.} & \vspace{0.5cm}

\textbf{For what purpose} \hlc[cyan!20]{do} \hlc[yellow!40]{South African women and children} \hlc[pink]{walk} \hlc[yellow!40]{a daily distance equivalent to 16 trips to the moon and back} \hlc[green!40]{collectively} ? \vspace{0.5cm}\\
Cause (\hlc[cyan!40]{CAU}) &
\textbf{Why} \hlc[cyan!20]{mainAux} \hlc[yellow!40]{nsubj} otherAux \hlc[pink]{verb} \hlc[yellow!40]{obj} \hlc[green!40]{modifiers} ? &
\hlc[cyan!40]{Since the average faucet releases 2 gallons of water per minute,} you can save up to four gallons of water every morning by turning off the tap while you brush your teeth. & 
\textbf{Why} \hlc[cyan!20]{can} \hlc[yellow!40]{you} \hlc[pink]{save} \hlc[yellow!40]{up to four gallons of water} \hlc[green!40]{by turning off the tap while you brush your teeth every morning} ? \vspace{0.5cm}\\
& & & \\
Temporal (\hlc[cyan!40]{TMP}) &
\textbf{When} \hlc[cyan!20]{mainAux} \hlc[yellow!40]{nsubj} otherAux \hlc[pink]{verb} \hlc[yellow!40]{obj} \hlc[green!40]{modifiers} ? \newline \vspace{0.5cm}
\textbf{Till when} \hlc[cyan!20]{mainAux} \hlc[yellow!40]{nsubj} otherAux \hlc[pink]{verb} \hlc[yellow!40]{obj} \hlc[green!40]{modifiers}? \newline \vspace{0.5cm} &
Stephen Hawking \hlc[cyan!40]{once on June 28, 2009} threw a party for time-travelers but he announced the party \hlc[cyan!40]{the next day.} \newline \vspace{0.5cm}
Princess Sita travelled the whole town \hlc[cyan!40]{until the end of summer.} 
&
\textbf{When} \hlc[cyan!20]{did} \hlc[yellow!40]{Stephen Hawking} \hlc[pink]{throw} \hlc[yellow!40]{a party} \hlc[green!40]{for time - travelers} ? 
\newline
\textbf{When} \hlc[cyan!20]{did} \hlc[yellow!40]{Stephen Hawking} \hlc[pink]{announce} \hlc[yellow!40]{the party} ? \vspace{0.5cm}
\newline
\textbf{Till when} \hlc[cyan!20]{did} \hlc[yellow!40]{Princess Sita} \hlc[pink]{travel} \hlc[yellow!40]{the whole town}? \\
& & & \\
Extent (\hlc[cyan!40]{EXT}) &
\textbf{By how much} \hlc[cyan!20]{mainAux} \hlc[yellow!40]{nsubj} otherAux \hlc[pink]{verb} \hlc[yellow!40]{obj} \hlc[green!40]{modifiers} ? &
New Mexico was won by Obama \hlc[cyan!40]{by a margin of 5\%} in 2008. & 
\textbf{By how much} \hlc[cyan!20]{was} \hlc[yellow!40]{New Mexico} \hlc[pink]{won} by \hlc[yellow!40]{Obama} \hlc[green!40]{in 2008}? \\
\hline
\end{tabular}
\caption{The templates of temporal, direction, extent, etc.\ are leveraged to ask questions about different modifiers. Answer fragments are highlighted in \hlc[cyan!40]{blue}. In passive cases like the last example, we change the template order from subj-verb-obj to obj-verb-by-subj.}
  \label{srl-table}
\end{table*}

We also use WordNet~\cite{miller1998wordnet} hypernyms of all potential short answers and replace \emph{\textbf{What}} with the bigram \emph{\textbf{Which hypernym}}. So, for a sentence like ``Hermione plays badminton at the venue'', we generate a question ``\emph{\textbf{Which sport}} does Hermione play at the venue?''. For computing the hypernym, we use the sense disambiguation implementation of~\citet{pywsd14}. While supersenses do display a richer lexical variety, sense definitions don't always fit well.

\subsection{Handling modals and auxilliaries}
During explicit inversion of the verb and arguments around it via our templates, we tried to ensure that the positions of auxiliaries are set, and negations are correctly treated. We define a few simple rules to ensure that. 
\begin{itemize}
\itemsep0em 
\item When there are multiple auxiliaries, we only invert the first auxiliary while the second and further auxiliaries remain as they are just before the main verb. 
\item We make the question auxiliary finite and agree with the subject. 
\item We ensure that the object is kept immediately after the verb. 
\item For passive cases, \textit{subj-verb-obj} is changed to \textit{obj-verb-by-subj}.
\end{itemize}

\subsection{Handling Factualness via Implicature} 
Previous rule-based approaches ~\cite{mazidi2016infusing, flor2018semantic} have used the NEG dependency label to identify polarity. But such an approach would suffer whenever polarities would be hierarchically entailed from their parent clauses in cases like ``Picard did not fail to X'' where the entailed polarity of ``X'' is, in fact, positive. Moreover, in one-way implications like ``Bojack hesitated to X'', it would be best not to generate a question for unsure cases since it is open-ended if Bojack did or did not X. A similar example is displayed in Figure~\ref{fig:PB3}.
For each verb representing a subordinate clause, we compute its entailed truth or falsity from its parent clause using the set of one-way and two-way implicative verbs, and verb-noun collocations provided by~\citet{karttunen2012simple}. For example, the two-way implicative construction ``forget to X'' entails that ``X'' did not happen, so it would be wrong to ask questions about ``X''.~\citet{karttunen2012simple} provides simple implications in the form of 92 verbs and phrasal implications in the form of 9 sets of verbs and 8 sets of nouns making 1002 verb-noun collocations. The entailed polarity of a clause can be either TRUE, FALSE, or UNSURE\footnote{Unsure clauses appear in one-way implicatives when it's unclear if the clause is true or false under either an affirmative or a negative parent clause.}. For FALSE clauses, we only generate a boolean question with a NO answer. For UNSURE clauses, we do not generate any question. For TRUE clauses and verbs and collocations not present in the above set, we rely on the NEG label.

\subsection{VerbNet Predicate Templates}
While SRL's event-based representations have permitted us to generate questions that talk about the roles participants of an event play, we exploit VerbNet's sub-event representation to ask questions on how participants' states change across the time frame of the event. In Figure~\ref{fig:vn4}, the event murder (VerbNet class \href{https://uvi.colorado.edu/verbnet/murder-42.1}{\textit{murder-42.1}}) results in a final state in which the participant \textit{Julius Caesar} is in a \textit{not-alive} state.

\begin{figure}
\begin{center}
\includegraphics[scale= 0.3]{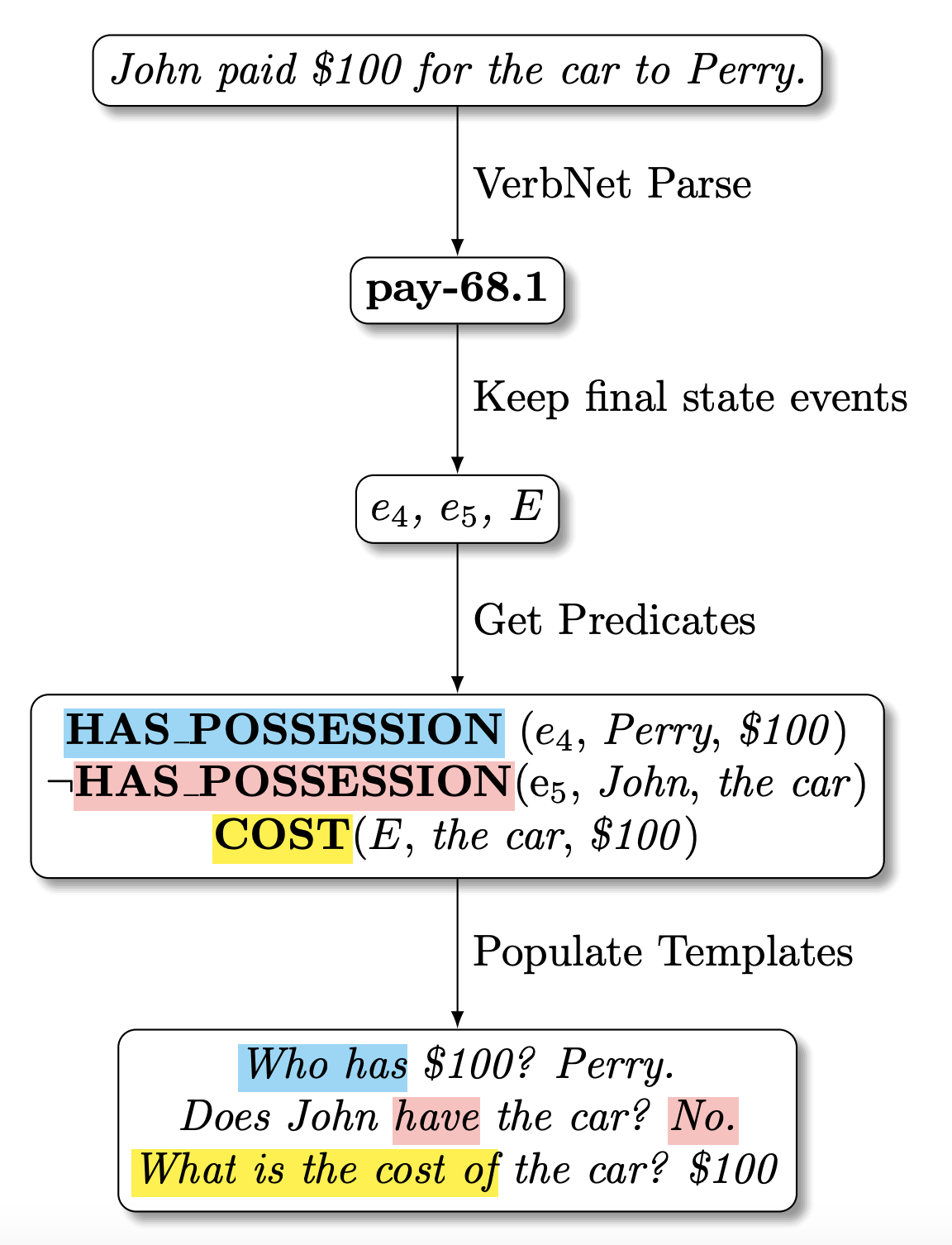}
\end{center}
  \caption{VerbNet Predicate Question Generation. All the predicates of the two sub-events $e_{4}$ and $e_{5}$ (HAS\_POSSESSION) would be considered since $e_{3}$ possesses a process-oriented predicate TRANSFER. COST is the predicate of the main event E.}
  \label{fig:vn1}
\end{figure}

Each class in VerbNet~\cite{schuler2005verbnet,brown2019verbnet} includes a set of member verbs, the thematic roles used in the predicate-argument structure, accompanied with flat syntactic patterns and their corresponding semantic predicates represented in neo-Davidsonian first-order-logic formulation. These semantic predicates bring forth a temporal sequencing of sub-events tracking how participants' states change over the course of the event. The advantage is to be able to ask questions bearing a surface form different from the source sentence but which are driven by reasoning rather than just being paraphrastic. For example, in the sentence, ``Brutus murdered Julius Caesar'', the event \href{https://uvi.colorado.edu/verbnet/murder-42.1}{\textit{murder-42.1}} entails a final state of ``death'' or the \textit{Patient} participant not being alive at the end of the event. So, we construct a template ``mainAux the Patient otherAux not alive?''. Similarly, the event \href{https://uvi.colorado.edu/verbnet/pay-68-1}{\textit{pay-68-1}} results in a final state in which the \textit{Recipient} ``Perry'' has possession of ``\$100'' and the \textit{Agent} ``John'' has possession of ``the car'', against which we define the templates as shown in Figure~\ref{fig:vn1}.

We formulate two sets of questions: boolean type and which-type questions asking specifically about these states. We create templates for VerbNet's stateful predicates like \textbf{has\_location, has\_possession, has\_information, seem, has\_state, cost, desire, harmed, has\_organization\_role, together, social\_interaction, authority\_relationship,} etc.\ which are present in 64.4\% of the member verbs in VerbNet\footnote{Out of 4854 member verbs, there are 3128 members whose syntactic frame contains at least one of these predicates.}. We outline a few of the templates in Table~\ref{vn-table}. 

During inference time, we first compute the VerbNet sense, the associated thematic role mapping, and syntactic frame (along with the predicates) with the help of~\citet{brown2019verbnet}'s parser. VerbNet's predicates are governed by the sub-events in which they occur. Although VerbNet's representation lays out a sequence of sub-events, no sub-event is explicitly mentioned as the final one\footnote{or a sub-event, which is an outcome of a process}. We choose all the predicates of those sub-events which are preceded by other sub-events which possess at least one process-oriented predicate.\footnote{Out of 174 VerbNet predicates, we manually categorize 84 predicates like HAS\_LOCATION, HAS\_POSSESSION as stateful predicates and the remaining ones like DESCRIBE, TRANSFER, etc. as process-oriented predicates.}

\begin{table*}[!hbpt]
\centering
\small
\begin{tabular}{| M{2.8cm} | L{3cm} | C{5.5cm} | R{3cm} |}
\hline
\textbf{Triggering Predicate and Thematic Arguments} &
\textbf{Question Template} &
\textbf{Sample Fact \& VerbNet Predicate} &
\textbf{Generated Question} \\
\hline
& & & \\
\textbf{HAS\_POSSESSION}  \newline (Asset,Recipient) &
\textbf{Who} has Asset ? Recipient &
Robert paid \hlc[pink]{\$100} to \hlc[cyan!40]{Mary} for the cycle. \newline
\textbf{HAS\_POSSESSION}(\hlc[cyan!40]{Mary},\hlc[pink]{\$100})
&
\textbf{Who} has \hlc[pink]{\$100} ? \hlc[cyan!40]{Mary}  \\
& & & \\
\textbf{HARMED}  \newline (Patient) &
\textbf{What} is harmed ? Patient &
The terrorists bombed \hlc[pink]{the building}. \newline
\textbf{HARMED}(\hlc[pink]{the building})
&
\textbf{What} is harmed ? \hlc[pink]{the building} \\
& & & \\
\textbf{NOT\_ALIVE}  \newline (Patient) &
Is Patient alive ?  No. &
According to epics, Vishnu killed \hlc[cyan!40]{the demon Kaitabh}. \newline
\textbf{NOT\_ALIVE} (\hlc[cyan!40]{the demon Kaitabh})
&
Is \hlc[cyan!40]{the demon Kaitabh} alive ?  No.  \\

\hline
\end{tabular}
\caption{VerbNet predicate templates (simplified) along with sample questions with the thematic roles highlighted. A question is created from the concept of ``being alive'' which is not synonymous with but is an outcome of ``killing''.}
\label{vn-table}
\end{table*}

\subsection{PropBank Argument Descriptions}
PropBank rolesets' course-grained annotation of verb-specific argument definitions (``killer'', ``payer'', etc.) to represent semantic roles offers robustly specific natural language descriptions to ask questions about the exact roles participants play. Nonetheless, not all descriptions are suitable to be utilized directly in rigid templates. So, we incorporate back-translation to 1) get rid of grammatical errors propagated from incorrect parsing and template restrictions, and 2) eliminate rarely used Prop-Bank descriptions and generate highly probable questions.

While previous work in rule-based QG has used SRL templates and WordNet senses to describe the roles arguments around a verb play, previous SRL templates have always been verb-agnostic, and we believe there is a great deal of potential in PropBank descriptions. Moreover, WordNet supersenses do not always give rise to acceptable questions. On manual evaluation, question relevance decreased after incorporating templates with WordNet supersenses. Instead, we make use of PropBank's verb-specific natural language argument descriptions to create an additional set of templates. VerbNet senses have a one-to-one mapping with PropBank rolesets via the SemLink project~\cite{palmer2009semlink}. We hence make use of ~\citet{brown2019verbnet}'s parser to find the appropriate PropBank roleset for a sentence.

\begin{figure}
  \includegraphics[scale= 0.28]{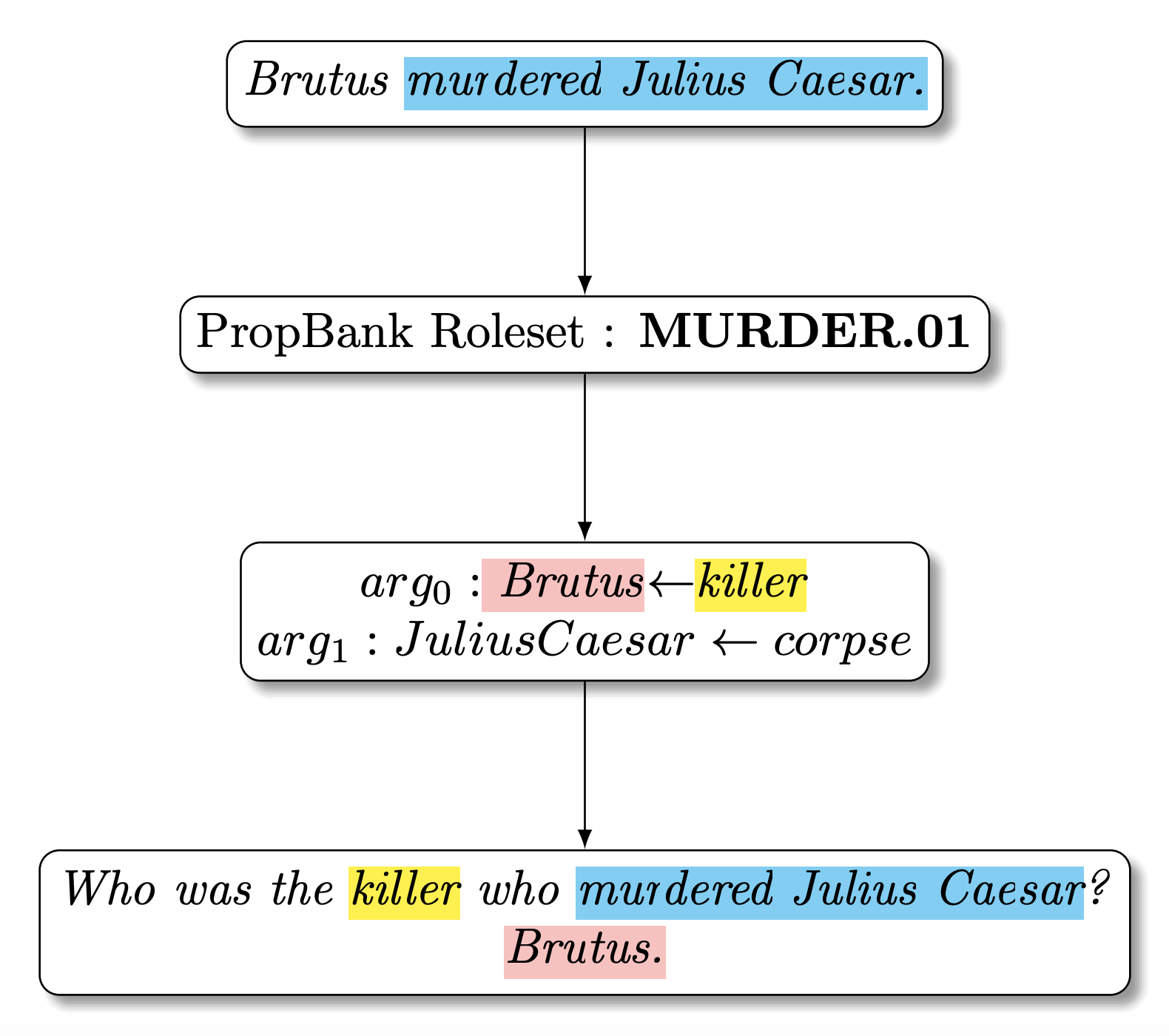}
  \caption{Here, ``killer'' is the natural language description of ``Brutus'' in the MURDER.01 roleset.}
  \label{fig:pb1}
\end{figure}

However, we observed that a lot of PropBank descriptions were noisy and made use of phrases which would be unarguably rare in ordinary parlance like ``breather'' or ``truster''. To eliminate such descriptions, we computed the mean Google N-gram probabilities~\cite{lin2012syntactic} of all the PropBank phrases in the timespan of the last 100 years and kept only those phrases which ranked in the top 50\%.  

\subsection{Back-Translation}
\begin{figure}
  \includegraphics[scale= 0.2]{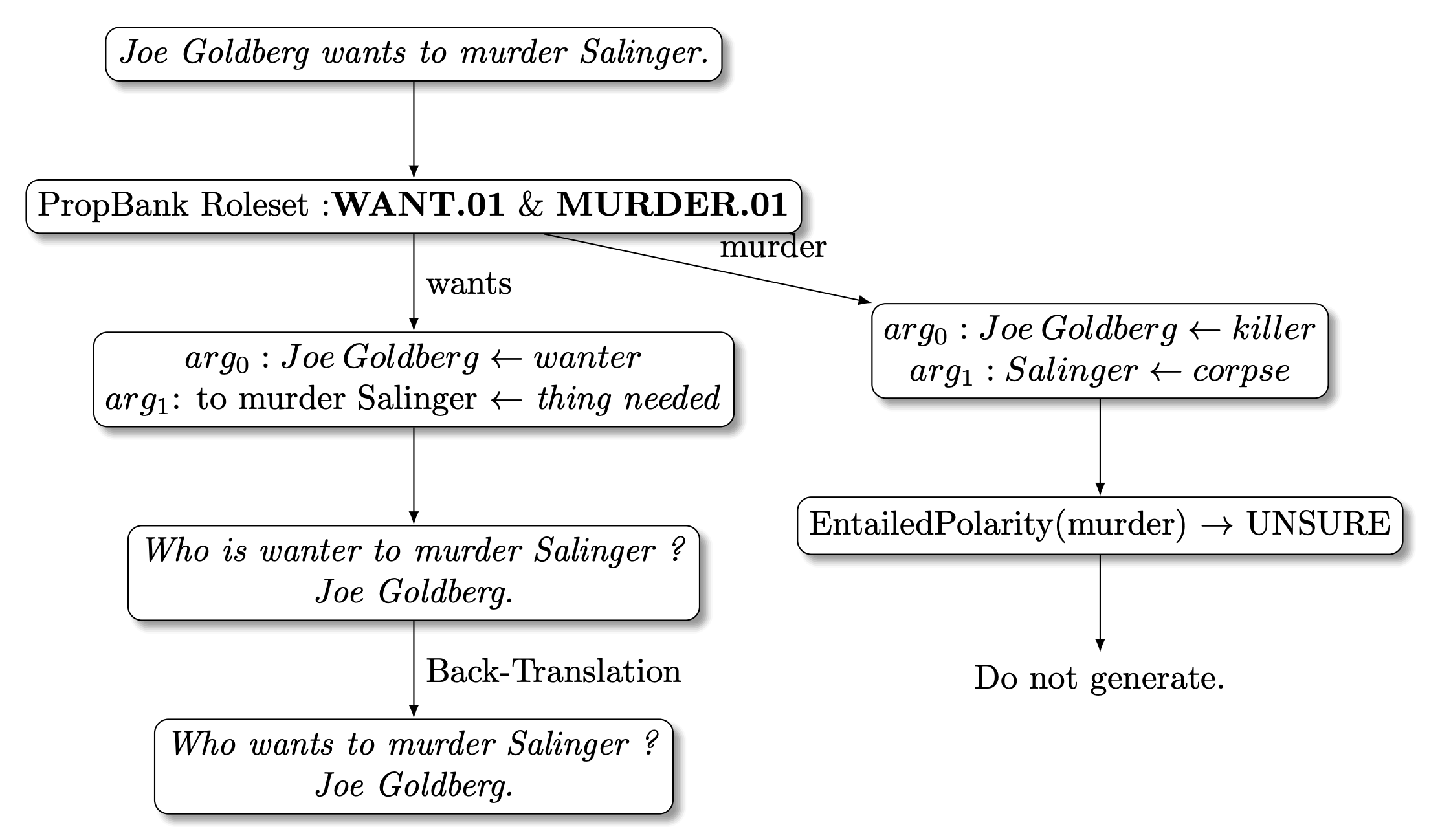}
  \caption{Back-translation and Implicature. Since the entailed polarity of ``murder'' is unsure, no questions are generated. }
  \label{fig:PB3}
\end{figure}

Back-translation has been used quite often in grammatical error correction~\cite{xie2018noising} and is well known to translate noisy and ungrammatical sentences to their cleaner high probability counterparts. We exploit this observation to clean questions with noisy and inconsistent PropBank descriptions like ``wanter'' (Figure~\ref{fig:PB3}). We use two state-of-the-art (SOTA) pre-trained transformer models \texttt{\small transformer.wmt19.en-de} and \texttt{\small transformer.wmt19.de-en} from \citet{ott2019fairseq} trained on the English-German and German-English translation tasks of WMT 2019. 

Figure~\ref{fig:synqg_all} in the Appendix shows the output of all the five sets of templates applied together over one sentence (along-with implicature).

\section{Evaluation and Results}
\begin{table*}[ht!]
\small
\centering
\begin{tabular}{| M{8.2cm} | L{1.4cm} | C{1.4cm} | R{1.4cm} | R{1.4cm} |}
\hline
\textbf{Architecture} & BLEU-1  & BLEU-2 & BLEU-3 & BLEU-4 \\
\hline
PCFG-Trans~\cite{heilman2010good} & 28.77 & 17.81 & 12.64 & 9.47 \\
SeqCopyNet~\cite{zhou2018sequential} &  &  &  & 13.02 \\
NQG++~\cite{zhou2017neural} & 42.36 & 26.33 & 18.46 & 13.51 \\
MPQG~\cite{song2017unified} &  &  &  & 13.91 \\
Answer-focused Position-aware model~\cite{sun2018answer} & 43.02 & 28.14 & 20.51 & 15.64 \\
To the Point Context~\cite{li2019improving} & 44.40 & 29.48 & 21.54 & 16.37 \\
s2sa-at-mp-gsa~\cite{zhao2018paragraph} & 44.51 & 29.07 & 21.06 & 15.82 \\
ASs2s~\cite{kim2019improving} & &  &  & 16.17 \\
CGC-QG~\cite{liu2019learning} & 46.58 & 30.9 & 22.82 & 17.55 \\
Capturing Greater Context~\cite{tuan2019capturing} & 46.60 & \textbf{31.94} & 23.44 & 17.76  \\
Natural QG with RL based Graph-to-Sequence~\cite{chen2019natural} & - & - & - & 17.94 \\
RefineNet~\cite{nema2019let} & \textbf{47.27} & 31.88 & 23.65 & 18.16 \\
QPP\&QAP~\cite{zhang2019addressing} & - & - & - & 18.37 \\
\hline
ACS-QG$^*$~\cite{liu2020asking} & \textbf{52.30}$^*$ & \textbf{36.70}$^*$ & \textbf{28.00}$^*$ & 22.05 \\
UNILM$^*$~\cite{wang2020minilm} & - & - & - & 24.32 \\
ERNIE-GEN$^*$~\cite{xiao2020ernie} & - & - & - & 25.57 \\
UNILMv2$^*$~\cite{bao2020unilmv2} & - & - & - & 26.30 \\
ProphetNet$^*$~\cite{yan2020prophetnet} & - & - & - & \textbf{26.72}$^*$ \\
\hline
%SynQG with VN/PB & 42.49  & 27.99 & 21.56 & 17.03 \\
Syn-QG & 45.55  & 30.24 & \textbf{23.84} & \textbf{18.72} \\
\hline
\end{tabular}
\caption{Automatic Evaluation Results on SQuAD of different QG models. PCFG-TRANS and Syn-QG are two rule-based models. *Work contemporaneous with or subsequent to the submission of this paper.}
\label{results-table}
\end{table*}

\begin{table*}[ht!]
\small
\centering
\begin{tabular}{| M{4.1cm} | L{3.4cm} | C{3.4cm} | R{3.4cm} |}
\hline
System & Avg. novel unigrams &
Avg. novel bigrams &
Avg. novel trigrams \\
\hline
H\&S & 23.6 & 40.64 & 52.22 \\
Syn-QG (w/o BT) & 26.8 & 43.93 & 53.4 \\
Syn-QG & 39.34 & 64.08 & 76.24 \\
SQUAD & 42.86 & 74.2 & 86.35 \\
\hline
Syn-QG (BT vs w/o-BT) & 28.78 & 55.18 & 67.81\\
\hline
\end{tabular}
\caption{The percentage of n-grams of the generated questions which are not present in the source sentence. The last row indicates the percentage of n-grams not present in the non-backtranslated questions.}
\label{vocab-dist-table}
\end{table*}

Most of the prior QG studies have evaluated the performance of the generated questions using automatic evaluation metrics used in the machine translation literature. We use the traditional BLEU scores~\cite{papineni2002bleu} and compare the performance of Syn-QG on the SQuAD~\cite{rajpurkar2016squad} test split created by \citet{zhou2017neural}. BLEU measures the average n-gram precision on a set of reference sentences. A question lexically and syntactically similar to a human question would score high on such n-gram metrics. Despite not utilizing any training data, Syn-QG performs better than the previous SOTA on two evaluation metrics BLEU-3 and BLEU-4 and close to SOTA on BLEU-1 and BLEU-2 (Table~\ref{results-table}) at the time of submission. The high scores obtained without conducting any training arguably shed a little light on the predictable nature of the SQuAD dataset too.

Besides SRL, Dependency, and NER templates, Syn-QG's questions also arise from VerbNet's predicates and PropBank's descriptions, which indeed by nature describe events not mentioned explicitly within the fact. Like in Figure~\ref{fig:vn1}, the sentence with the event ``paid'' results in a question with a stateful event of ``cost''. Deducible questions like these have a good chance of having a distribution of n-grams quite different from the source sentences, possibly exposing the weakness of traditional n-gram metrics and rendering them less useful for a task like QG.

In order to have a complete and more reliable evaluation to gauge the system, we also carry out a human evaluation using two of the metrics used in QG-STEC Task B~\cite{rus2012detailed}, namely grammaticality, and relevance which we define below. We compared the questions generated from our system against the constituency-based H\&S~\cite{heilmanquestion}, a neural system NQG~\cite{du2017learning} which does not depend on a separate answer extractor and QPP\&QAP\footnote{Since the QPP\&QAP model does not have a separate answer extractor, we use the answer spans computed from Syn-QG (412 in total after discarding overlaps).}~\cite{zhang2019addressing} which has outperformed existing methods. We fed a total of 100 facts randomly picked from Wikipedia and 5 commercial domains (IT, Healthcare, Sports, Banking and Politics) combined, to each of the four systems. We then conducted a crowd-sourced evaluation over Amazon Mechanical Turk for the generated questions.

\begin{itemize}
\item \textbf{Grammatical Correctness}: Raters had to rate a question on how grammatically correct it is or how syntactically fluent it is, disregarding its underlying meaning. 
\item \textbf{Relevance Score}: Raters had to give a score on how relevant the generated question is to the given fact. The relevance score helps us gauge whether the question should have been generated or not irrespective of its grammaticality.\footnote{In cases when the grammaticality is extremely low like 1 or 2, the relevance score will also tend to be low. Otherwise, we assume that minor grammatical variations can be ignored while gauging relevance.}
\end{itemize}
Each question was evaluated by three people scoring grammaticality and relevance on a 5 point Likert scale. The inter-rater agreement (Krippendorff's co-efficient) among human evaluations was 0.72. The instructions given to the Mturk raters are provided in the Appendix Figure~\ref{fig:mturk}. Syn-QG generates a larger number of questions than H\&S and performs strongly on grammaticality ratings. Syn-QG is also able to generate highly relevant questions without the use of a ranker. Also, rule-based approaches seem to be much better at generating relevant questions than neural ones.

QG-STEC also used variety and question types as their evaluation criteria and rewarded systems to generate questions meeting a range of specific question types. Syn-QG's questions cover each of those question types. 

Since many times, despite the ability to paraphrase (Table~\ref{vocab-dist-table}), back-translated outputs tend to change the meaning of the original sentence, we also measured back-translation's impact on the above QG metrics. While neural models are learning syntactic structures well, there is still some progress to be made to generate relevant questions.

\section{Discussion} 
We introduced Syn-QG, a set of broad coverage rules leveraging event-based and sub-event based sentence views along with verb-specific argument descriptions. Automatic and manual evaluations show that Syn-QG is able to generate a large number of diverse and highly relevant questions with better fluency. Verb-focused rules help approach long-distance dependencies and reduce the need for explicit sentence simplification by breaking down a sentence into clauses while custom rules like implications serve a purpose similar to a re-ranker to discard irrelevant questions but with increased determinism. 
While our work focuses on sentence-level QG, it would be interesting to see how questions generated from VerbNet predicates would have an impact on multi-sentence or passage level QG, where the verb-agnostic states of the participants would change as a function of multiple verbs. 
The larger goal of QG is currently far from being solved. Understanding abstract representations, leveraging world knowledge, and reasoning about them is crucial. However, we believe that with an extensible and transparent architecture, it is very much possible to keep improving the system continuously in order to achieve this larger goal.

\section*{Acknowledgments}
We thank the three anonymous reviewers for their
helpful comments and invaluable suggestions. We also thank the members of Amelia Science, RnD IPsoft, India - Manjunath Hegde, Anant Khandelwal, Ashish Shrivastava for their work in QG and especially Viswa Teja Ravi, for helping in replicating ~\citet{mazidi2016infusing}'s work. We also thank Uday Chinta and IPsoft, India, for supporting and providing access to Amazon Mechanical Turk.

\bibliography{anthology,acl2020}
\bibliographystyle{acl_natbib}

\appendix
%\iffalse % Appendix is disabled 
\section{Appendices}
\label{sec:appendix}

\begin{algorithm}[!hbpt]
\caption{Dependency Heuristics}
  \label{alg:the_alg}
\begin{algorithmic}
\State $\{d_0 \ldots d_n \} \gets dependency(w_0 \ldots w_n)$
\State \emph{loop i = 0, until i = n}:
\If {$\textit{parent}(d_i) != \textit{null}$}
\State $d_v \gets parent(d_i) $
\State $\{A_0 \ldots A_{CAU}\} \gets SRL(d_v)$
\State $subj \gets A_0$
\If {$d_i \in A_1$}
\State $obj \gets A_1$
\Else
\State $obj \gets A_2$
\EndIf
\State $A_{x} \gets \sum (A_3,A_4,... A_{TMP})$
\State $verb \gets \{d_v, modals, negation\}$
\State $template \gets dep_{type} \gets d_i$
\State $QA \gets template(subj, obj, verb, A_{x}) $
\State \textbf{close};
\EndIf
\end{algorithmic}
\end{algorithm}

\begin{algorithm}[ht!]
\caption{Named Entity Heuristics}\label{ner-algo}
\begin{algorithmic}
\State $\{SRL_1 \ldots SRL_s \} \gets SRL(w_0 \ldots w_n)$
\BState \emph{loop j = 0, until j = s}:
\If {$SRL_j \; contains \:A_0 \text{ or } A_1\: and\: at\: least\: 1\: A_m$}
\State $\{A_0 \ldots A_{CAU}, A_{TMP}\} \gets SRL_j$
\BState \emph{$loop \text{ } A_x \in SRL_j \; if \; A_x\;contains\;a\;NE$}:
\State $subj \gets A_0$
\State $A_{x}^{-} \gets \sum (A_3,A_4,... A_{TMP} - A_x)$
\State $verb \gets \{A_v, modals, negation\}$
\State $template \gets NE_{type} \gets A_x$
\State $QA \gets template(subj, A_x, verb, A_{x}^{-}) $
\State \textbf{close};
\EndIf
\end{algorithmic}
\end{algorithm}

\begin{table*}[!hbpt]
\small
\centering
\begin{tabular}{| M{2cm} | L{3cm} | C{5.7cm} | R{3.6cm} |}
\hline
\textbf{Potential Short Answer} \newline \textbf{(Named Entities)} &
\textbf{Question Template} &
\textbf{Sample Fact} &
\textbf{Generated Question} \\
\hline
Location &
\textbf{Where} mainAux subj otherAux verb obj modifiers ?  \vspace{0.5cm} &
\hlc[pink]{The event was organized } \hlc[cyan!40]{at Times Square.} & 
\textbf{Where} was \hlc[pink]{the event organized}?
 \\
 Person &
\textbf{Who} mainAux subj otherAux verb obj modifiers ? \textbf{Whom} mainAux obj otherAux verb modifiers   \vspace{0.5cm} &
\hlc[pink]{WestWorld brought back the life of}\hlc[cyan!40]{ the roboticist Craig Smith.} & 
\textbf{Whom} did \hlc[pink]{WestWorld bring back the life of}?
 \\
  Date &
\textbf{When} mainAux subj otherAux verb obj modifiers ?  \vspace{0.5cm} &
\hlc[pink]{Donald Trump won the elections} \hlc[cyan!40]{in the year 2016} & 
\textbf{When} did \hlc[pink]{Donald Trump win the elections}?
 \\
   Number &
\textbf{How many} mainAux subj otherAux verb obj modifiers?  \vspace{0.5cm} &
\hlc[cyan!40]{A thousand} \hlc[pink]{will not be enough for the event.} &
\textbf{How many} \hlc[pink]{will not be enough for the event}?
 \\
    Phone Number &
\textbf{At what number} mainAux subj otherAux verb obj modifiers ?  \vspace{0.1cm} &
\hlc[pink]{The pizza guy can be reached} \hlc[cyan!40]{at +91-748-728-781} &
\textbf{At what phone number} \hlc[pink]{can the pizza guy be reached}?
 \\
 Duration & \textbf{For how long} mainAux subj otherAux verb obj modifiers?  \vspace{0.1cm}
& \hlc[pink]{Lauren would be staying in the hut} \hlc[cyan!40]{for around 10 minutes.}
& \textbf{For how long} \hlc[pink]{would Lauren be staying at the hut}? 
 \\
 Organization & \textbf{Which organization} mainAux subj otherAux verb obj modifiers?  \vspace{0.1cm}
& \hlc[pink]{Deepak joined} \hlc[cyan!40]{the big firm, the United Nations.}
& \textbf{Which organization} \hlc[pink]{did Deepak join}? 
 \\
\hline
\end{tabular}
\caption{ SRL arguments which contain a named entity are fully considered as a short answer \hlc[cyan!40]{``for around 10 minutes''} rather than only the named entity span ``10 minutes''. SRL arguments are highlighted in \hlc[cyan!40]{blue}.
  }
\label{ner-table}
\end{table*}

\begin{figure*}%[ht!]
\begin{center}
  \includegraphics[width=\textwidth,height=\textheight,keepaspectratio]{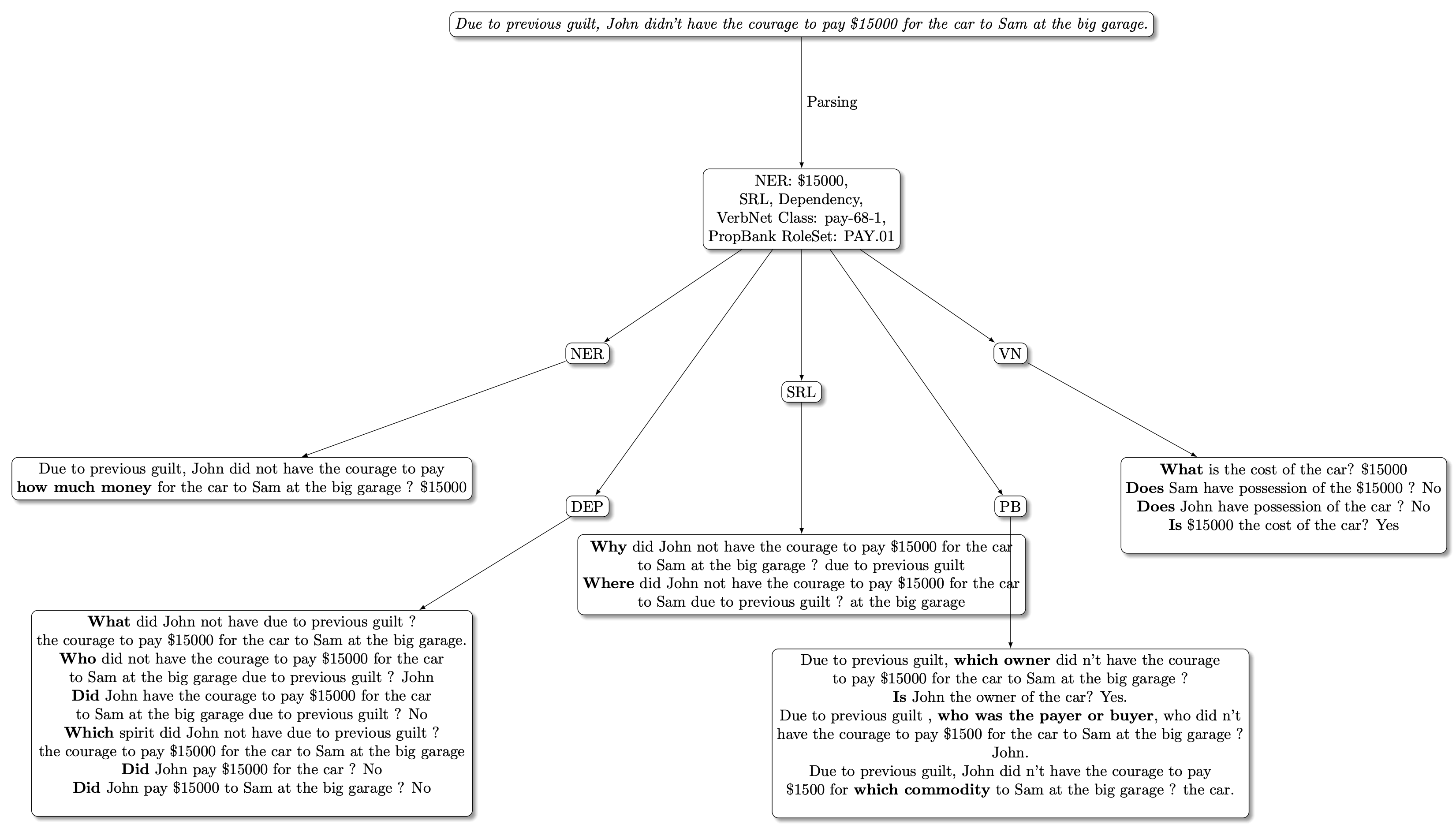}
  \end{center}
  \caption{Questions generated by each set of heuristics for one sentence which are further sent for back-translation.}
  \label{fig:synqg_all}
\end{figure*}
\begin{figure*}[t]
  \includegraphics[width=\textwidth,height=\textheight,keepaspectratio]{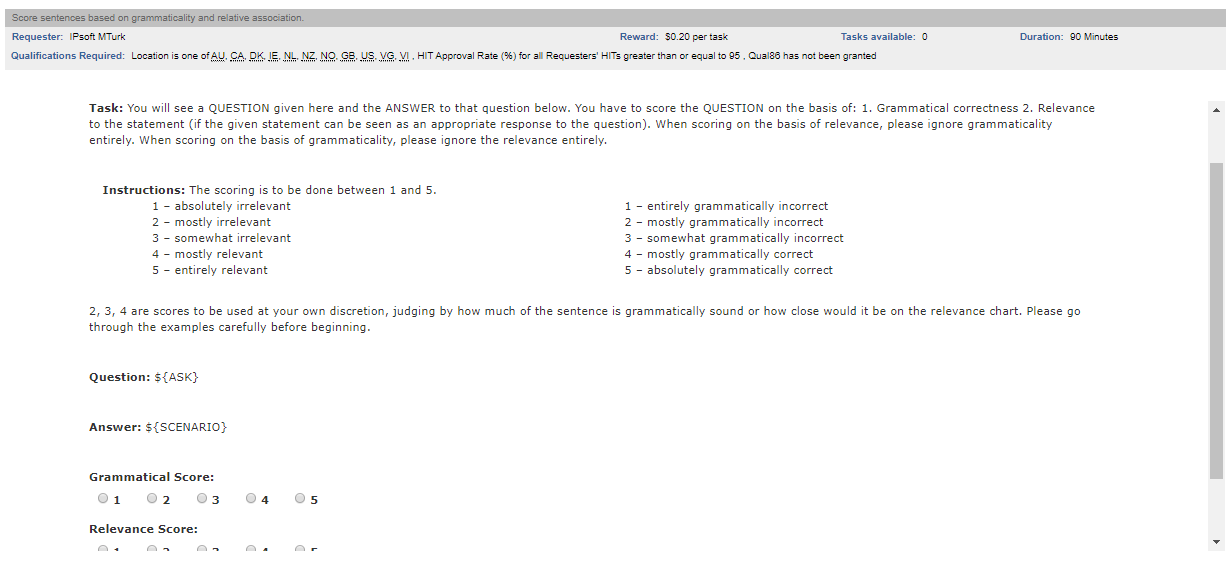}
  \caption{The MTURK template used for collecting responses for measuring question relevance and grammaticality.}
  \label{fig:mturk}
\end{figure*}

% \begin{figure}[!ht
%   \includegraphics[width=\linewidth]{VN.png}
%   \caption{VerbNet Parse output of Brown's parser for ``John paid \$100 for the car to Perry''. Notice the predicates in the final sub-event ``e5'' and main event ``E''}
%   \label{fig:vn_ui}
% \end{figure}

%\section{Supplemental Material}
%\label{sec:supplemental}
\fi
\end{document}